# A light-weight method to foster the (Grad)CAM interpretability and explainability of classification networks

Alfred Schöttl
*Dept. of Electrical Engineering and Information Technology*
*University of Applied Sciences Munich*
Munich, Germany
alfred.schoettl@hm.edu

*Abstract*—We consider a light-weight method which allows to improve the explainability of localized classification networks. The method considers (Grad)CAM maps during the training process by modification of the training loss and does not require additional structural elements. It is demonstrated that the (Grad)CAM interpretability, as measured by several indicators, can be improved in this way. Since the method shall be applicable on embedded systems and on standard deeper architectures, it essentially takes advantage of second order derivatives during the training and does not require additional model layers.

## I. INTRODUCTION

The training of possibly weak features and the improved interpretability of the obtained result by self-supervised reinforcing is a recent problem in the design of localized classification networks.

[1] propose the idea of self-attention distillation (SAD) and show that the training process of weak features can be supported in a self-learning way. SAD is a self-supervised method which exploits attention maps extracted from the network during training. Every convolutional layer is enriched by an additional attention layer. An appropriate extension of the loss function guarantees the consistency of the attention maps between the different layers. In this way, also the ordinary layers are enforced to build strong feature detectors early during the training.

SAD shows its performance in the detection of weak features with extended and directional shapes such as lane markings and is especially designed for networks with few layers. In the sequel, we propose a method inspired by SAD to strengthen the representation of deeper networks and to improve the explainability of the network.

With the success of deep neural networks, the need arose to explain why a network works well in a particular application. Explainability methods aim at gaining deeper insights for this reasoning [2]. As an important subclass of models we consider the class of localizing networks (such as convolutional networks) with softmax output trained to classify samples containing singular objects of these classes.

GradCAM [3] and related methods (including class activation maps (CAM) [11], GradCAM++ [10] etc.) are designed to explain how the classification works by localizing regions of activation. They are model agnostic and in the framework of [1] characterized as sensibility-based methods. These methods give considerable insights into the localization of activations in a specific network layer of a trained network and are proven to work well in practice.

While CAM requires the presence of a global pooling layer after the convolutional layers, GradCAM overcomes this restriction by using gradients of the class scores with respect to the activation maps as weights which are combined with the activation maps itself to obtain a localized map of attentions which are responsible for the decision for the specific class.

Our method is designed to encourage localized GradCAM activation maps. The interpretability of the CAMs even for weaker features is fostered by pushing the gradient in a direction that produces more distinct class activation maps. Since the method shall be applicable on embedded systems and on standard deeper architectures such as Resnet [5], it essentially takes advantage of second order derivatives during the training and does not require additional model layers which would lead to an increased inference time.

The main contribution of this work is the concept to modify the training loss by a term based on novel GradCAM-related measures which evaluate the explainability of localized classification networks without adding additional layers to the network. It is demonstrated that various measures addressing different aspects of the CAM interpretability are concurrently improved. The price of the interpretability improvement is a moderate performance degradation. The balance between these two targets can be adjusted by a hyperparameter.

## II. RELATED WORK

Apart from the CAM methods, a lot of other features and methods such as the SHAP family (Shapely Additive explanations, [6]) have recently been proposed to give insights into the classification performance of convolutional nets.

SAD methods [1] improve the classification performance and the interpretability by the introduction of a self-supervised attention mechanism.

The probably most closely related and highly effective approach are Attention Branch Networks (ABN) [7] which add layers in attention branches in the network reproducing and optimizing CAM features. In this way, CAM features are not only used for visualization but also to optimize the interpretability of the network.

## III. METHODS

### A. CAM Interpretability Measures

Interpretability is often defined as the ability of a human to understand the result a neural network outputs [8]. A key point in the context of localized networks such as convolutional networks is the ability to separate contents from attention. This kind of partitioning allows to investigate which regions and which contents in these regions are responsible for the classification result. Explainability methods in this context shall produce a map of relevant locations given an input, the network structure and the resulting output.

We hypothesize that a decision is the better interpretable the more localized and sharp the map of the true class $c$ is. A (Grad)CAM map $M$ in which all feature locations take equally part to the result does not give much information (except the pure existence of the feature), whereas a single activated feature pixel is a strong explanation. We therefore propose three measures to quantify the local CAM interpretability:

- The *CAM entropy ce*,

$$ce(M) := -\sum \bar{M}_{ij} \ln \bar{M}_{ij}, \qquad (1)$$

with the normalized map $\bar{M}_{ij} = \frac{M_{ij}}{\sum M_{kl}}$, and the map pixel indices $ij$ and $kl$ respectively,

- The *CAM ellipsoidal area ca*,

$$ca(M) := \sqrt{\lambda_1 \lambda_2}, \qquad (2)$$

with $\lambda_k$ being the eigenvalues of the covariance matrix of the two-dimensional probability function defined by $\bar{M}$, and

- The *CAM dispersion cd*,

$$cd(M) := \frac{\sigma_M^2}{\mu_M^2} \qquad (3)$$

for the variance $\sigma_M^2$ and the mean value $\mu_M$ of all CAM values in (the flattened) $M$.

The three measures address fundamentally different properties. The entropy measures the amount of activated areas (independent of the neighborhoods), the ellipsoidal area measures the concentration of the activated area, and the dispersion the uniqueness of the decision. A lower CAM entropy, a lower CAM ellipsoidal area and a higher CAM dispersion are indicators of more distinct and sharper CAMs.

The CAM dispersion is an important indicator for the uniqueness of the decision. Observe that in classical GradCAM visualizations the activation values are rescaled and offset shifted [3],

$$M_{visual} = \frac{M - \min(M)}{\max(M) - \min(M)}.$$

This normalization is a crucial point in the interpretation process of (Grad)CAMs: Since the minimum is not known in advance, it is needed to obtain the required visual representation. On the other hand, the rescaling masks gradual differences in $M$ and may suggest a distinct decision of the net even though the CAM dispersion is very low.

### B. Network Loss and Training

We assume a classical convolution classification network consisting of a stack of convolutional layers with activations $A^k = \left(A_{ij}^k\right)_{ij}$ and a stack of subsequent dense layers with output logits vector $y$ (in front of the softmax activation). The network is trained by optimizing the cross-entropy loss function $l_{\text{train}} = l_{\text{train}}(y_{\text{true}}, y_{\text{pred}})$.

Gradient activation maps for a specific class $c$ are computed by [3]

$$L_{GCAM}^c = \text{relu}(\sum_k \alpha_k^c A^k). \qquad (4)$$

The weights $\alpha_k^c$ are obtained by global average pooled gradients of the class score (the inference result before the softmax) with respect to the activations,

$$\alpha_k^c = \frac{1}{Z} \sum_{ij} \frac{\partial y^c}{\partial A_{ij}^k}. \qquad (5)$$

with a constant $Z$. Observe that the necessary gradients can be computed by standard backprop.

To foster the explainability of the model and the sharpness of Grad-CAM maps during training, we modify the training loss $l_{train}$ with an entropy loss term (remember that the CAM entropy requires the normalized map) by

$$l = l_{\text{train}} + \beta\, ce(L_{GCAM}^c) \qquad (6)$$

It is remarkable that application of the usual optimization algorithms such as gradient descent or Adam now leads to a second order method.

It turns out that the CAM entropy reduces the size of the active area in the activation maps and sharpens the maps. The ordinary loss function maximizes the classification accuracy and hence prevents the active area from tattering or infinitely shrinking. The more complex interpretability measures $ca$ and $cd$ could also be used here. Experiments show that the optimization results with the measures $ca$ and $cd$ are only slightly better than the results obtained by using $ce$ at the price of significantly greater computational effort. The results (see next section) indicate that optimizing one of these three measures also optimizes the others.

We could also encourage the maps of the other classes to produce weaker results by adding the negative loss terms if it is known that only the true class is present in the sample.

The hyper-parameter $\beta$ is crucial for a well-adjusted performance of the model. While $\beta = 0$ gives the best accuracy, a greater value of $\beta$ results in more localized and better explainable models. In practice, the performance drop at moderate $\beta$-values is small.

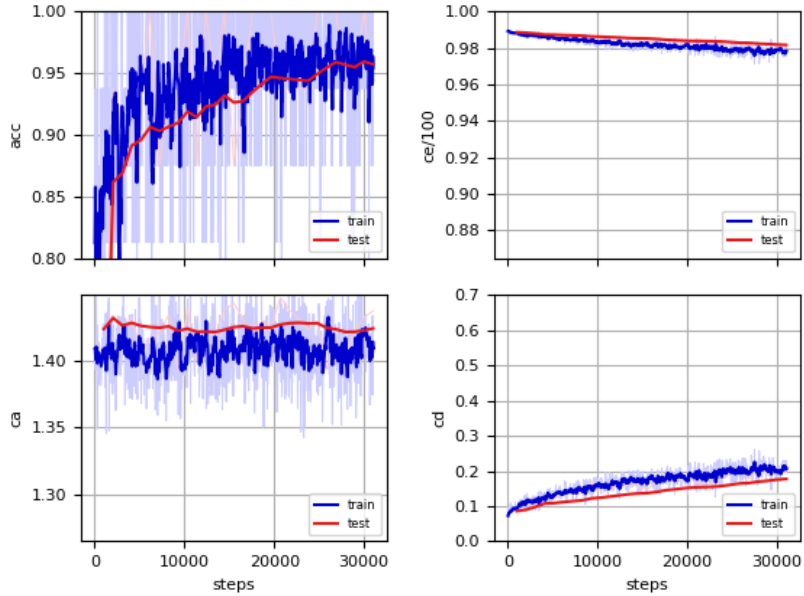

Fig. 2. Accuracy (upper left), CAM entropy (upper right), CAM ellipsoidal area (lower left) and CAM dispersion (lower right) of the network trained with $\beta = 0$ for training (blue) and test (red) data. The x-axes show the training steps. Best viewed in color.

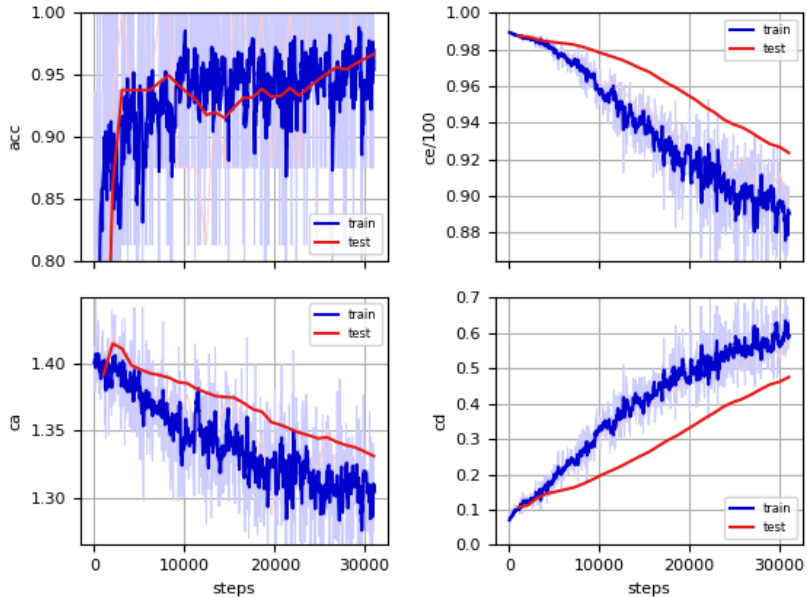

Fig. 1. Accuracy (upper left), CAM entropy (upper right), CAM ellipsoidal area (lower left) and CAM dispersion (lower right) of the network trained with $\beta = 100$ for training (blue) and test (red) data. The x-axes show the training steps. Best viewed in color.

## IV. EXPERIMENTS

### A. Setup

We use the convolutional part of the Resnet50 architecture [5] with its 50 layers and 23.5 million weights pretrained on imagenet as testbed. Two dense layers are added to the average pooled and flattened output. The first dense layer with 300 outputs is endowed with relu activation functions and the second with a softmax.

We use the boxes of the PASCAL VOC 2012 dataset [9], resized at 192x192 pixels, as training and test data. The dataset contains 20 object classes. Since there may be several object boxes, we draw for each epoch and sample one box at random from each sample. The object classes are significantly unbalanced, no measure is taken to rebalance or re-weight the data. There are 16551 training samples and 4952 test samples with up to 12 objects in the dataset.

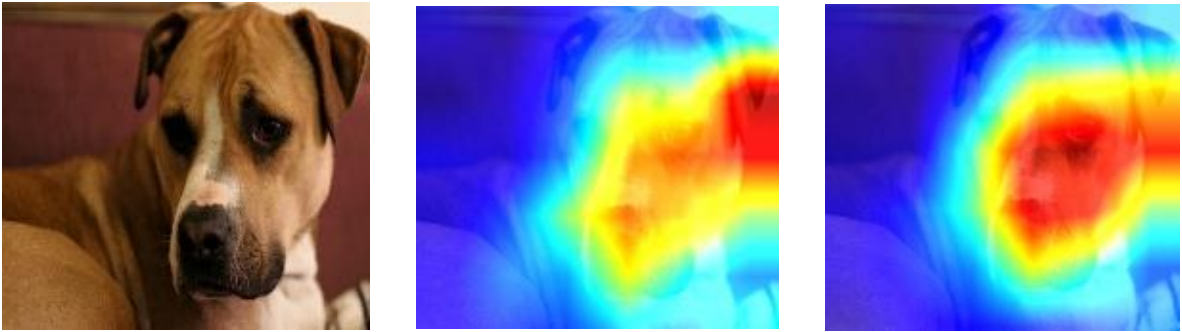

Fig. 3. An exemplary sample of the class "dog" and the GradCAM visualizations for the trained network with $\beta = 0$ (middle) and $\beta = 100$ (right). The GradCAM visualization normalizes the map. The difference between the blue low activated and red high activated regions is 0.33 in the middle image and 1.18 in the right image. This corresponds to a relative difference in the maps of 32% and 114%. Best viewed in color.

The network is trained for 30 epochs with the Adam optimizer at a constant learning rate of 0.001. The used loss function is given by equation (6). A batch size of 20 is chosen. All three interpretability figures (1)-(3) are collected for the training and the test data during training at reasonable frequencies.

*B. Results*

Fig. 1 shows the learning curve of the training and the test accuracy of the regularly trained network ( $\beta = 0$ ). Furthermore, the interpretability figures are depicted. While the ellipsoidal area almost remains constant after the very short initialization phase, the entropy slightly decreases. The most significant change is visible in the dispersion which enlarges from 0.1 to 0.2. We can therefore state that the activation areas are settled very early. The further training adapts the weights between features at these locations and the classification logits.

We then repeat the training with a loss function extended by the CAM entropy *ce* as an additive term (with a large $\beta = 100$ for demonstration purposes). Fig. 2 depicts a significantly different situation. A part of the learning effort is now put into an improved CAM interpretability, the training and the test accuracy curves are shallower and exhibit a slightly worse performance. The mean ellipsoidal activation area is reduced by 20 % and the dispersion is enlarged by the factor 6. This training therefore continuously adapts regions and the associated contents processing. The networks learns to watch more precisely and to decide with more determination.

TABLE I. CHARACTERISTICS OF AN EXEMPLARY SAMPLE

|                   | $\beta = 0$ | $\beta = 100$ |
|-------------------|-------------|---------------|
| class probability | 0.999       | 0.999         |
| max logit         | 20.1        | 80.8          |
| min logit         | -40.7       | -4.9          |
| max GradCAM       | 1.358       | 2.220         |
| min GradCAM       | 1.029       | 1.039         |
| ce                | 0.991       | 0.923         |
| ca                | 1.386       | 1.249         |
| cd                | 0.077       | 0.257         |

Fig. 3 depicts a sample of the test data set and the GradCAM visualization in both cases. Observe that the activated area in the case $\beta = 100$ is significantly smaller and sharper. There are considerable differences in the GradCAMs as summarized in Tab. 1. Observe that the dispersion (and hence the sensitivity between the activated and the irrelevant region) is substantially higher. The relative difference between inactivated and activated regions is 33% for $\beta = 0$ and 114% for $\beta = 100$. The net focuses more on the most meaningful parts (in this case the snout and the eyes).

The computation was performed on an nVidia 1080Ti graphics card with Tensorflow2. One epoch took (including test step and logging) with the conventional loss 4:35 min and with the loss extension 8:22 min.

## V. CONCLUSION

We investigated the possibility to foster the CAM interpretability of a network purely by modification of the loss function. For this purpose, we introduced three GradCAM-based interpretability measures and used the GradCAM entropy as additional loss term. An experiment utilizing Resnet50 and PASCAL VOC data demonstrated the feasibility.